\documentclass[lettersize,journal]{IEEEtran}
\usepackage{amsmath,amsfonts}
\usepackage{algorithm}
\usepackage{array}
\usepackage[caption=false,font=normalsize,labelfont=sf,textfont=sf]{subfig}
\usepackage{textcomp}
\usepackage{stfloats}
\usepackage{url}
\usepackage{verbatim}
\usepackage{graphicx}
\usepackage{cite}
\usepackage{graphicx}%
\usepackage{multirow}%
\usepackage{amsmath,amssymb,amsfonts}%
\usepackage{amsthm}%
\usepackage{mathrsfs}%
\usepackage{xcolor}%
\usepackage{textcomp}%
\usepackage{manyfoot}%
\usepackage{booktabs}%
\usepackage{algorithm}%
\usepackage{algorithmicx}%
\usepackage{algpseudocode}%
\usepackage{listings}%
\usepackage{graphicx}
\usepackage{multirow}
\usepackage{caption}
\usepackage{subcaption}

\usepackage{stackengine}

\usepackage{pifont}% http://ctan.org/pkg/pifont
\newcommand{\cmark}{\ding{51}}%
\newcommand{\xmark}{\ding{55}}%

\begin{document}

\title{XPoint: A Self-Supervised Visual-State-Space based Architecture for Multispectral Image Registration}

\author{Ismail Can Yagmur, Hasan F. Ates, Bahadir K. Gunturk
\thanks{I. C. Yagmur and H. F. Ates are with the Faculty of Engineering, Ozyegin University, Istanbul, Turkey (e-mails: can.yagmur@ozu.edu.tr, hasan.ates@ozyegin.edu.tr).}
\thanks{Bahadir K. Gunturk is with the School of Engineering and Natural Sciences, Istanbul Medipol University, Istanbul, Turkey (e-mail: bkgunturk@medipol.edu.tr).}
}

% \author{İsmail Can Yağmur,~\IEEEmembership{Student Member,~IEEE}, Hasan F. Ates,~\IEEEmembership{Member,~IEEE}
% \thanks{İsmail Can Yağmur is with the Department of Computer Science, Ozyegin University, Istanbul, Turkey (e-mail: ismail.yagmur@ozu.edu.tr).}
% \thanks{Hasan F. Ates is with the Faculty of Engineering, Ozyegin University, Istanbul, Turkey (e-mail: hasan.ates@ozu.edu.tr).}
% }

        % <-this % stops a space



\maketitle

\begin{abstract}
Accurate multispectral image matching presents significant challenges due to non-linear intensity variations across spectral modalities, extreme viewpoint changes, and the scarcity of labeled datasets. Current state-of-the-art methods are typically specialized for a single spectral difference, such as visible-infrared, and struggle to adapt to other modalities due to their reliance on expensive supervision, such as depth maps or camera poses. To address the need for rapid adaptation across modalities, we introduce XPoint, a self-supervised, modular image-matching framework designed for adaptive training and fine-tuning on aligned multispectral datasets, allowing users to customize key components based on their specific tasks. XPoint employs modularity and self-supervision to allow for the adjustment of elements such as the base detector, which generates pseudo-ground truth keypoints invariant to viewpoint and spectrum variations. The framework integrates a VMamba encoder, pre-trained on segmentation tasks, for robust feature extraction, and includes three joint decoder heads: two are dedicated to interest point and descriptor extraction; and a task-specific homography regression head imposes geometric constraints for superior performance in tasks like image registration. This flexible architecture enables quick adaptation to a wide range of modalities, demonstrated by training on Optical-Thermal data and fine-tuning on settings such as visual-near infrared(0.75–1.4 $\mu$m), visual-infrared(3-8 $\mu$m), visual-longwave infrared(0.8–15 $\mu$m), and visual-synthetic aperture radar. Experimental results show that XPoint consistently outperforms or matches state-of-the-art methods in feature matching and image registration tasks across five distinct multispectral datasets. Our source code is available at https://github.com/canyagmur/XPoint.
\end{abstract}


\maketitle

\section{Introduction}
\label{sec:intro}
Image matching \cite{imagematching1} identifies and correlates similar structures across multiple images, often by geometrically aligning a moving image with a fixed reference image in a process known as image registration. This technique plays a crucial role in applications such as remote sensing \cite{remotesensing}, visual localization \cite{visloc1}, homography estimation \cite{homography1}, and structure from motion (SfM) \cite{sfm1}. Multimodal image matching (MMIM) extends this concept by integrating data from different imaging modalities, allowing systems to interpret complex environments more effectively. This integration combines images captured by various sensors or under diverse conditions (e.g., different times of day \cite{daynight1}, weather \cite{crossweather1}, or seasons \cite{crossseason3}). MMIM also merges data types, such as combining images with textual information \cite{imagetext2}, capitalizing each modality’s unique strengths to provide a richer, more comprehensive understanding of the scene \cite{multimodalreview}.

\begin{figure*}[t]
    \centering
    \includegraphics[width=1\textwidth]{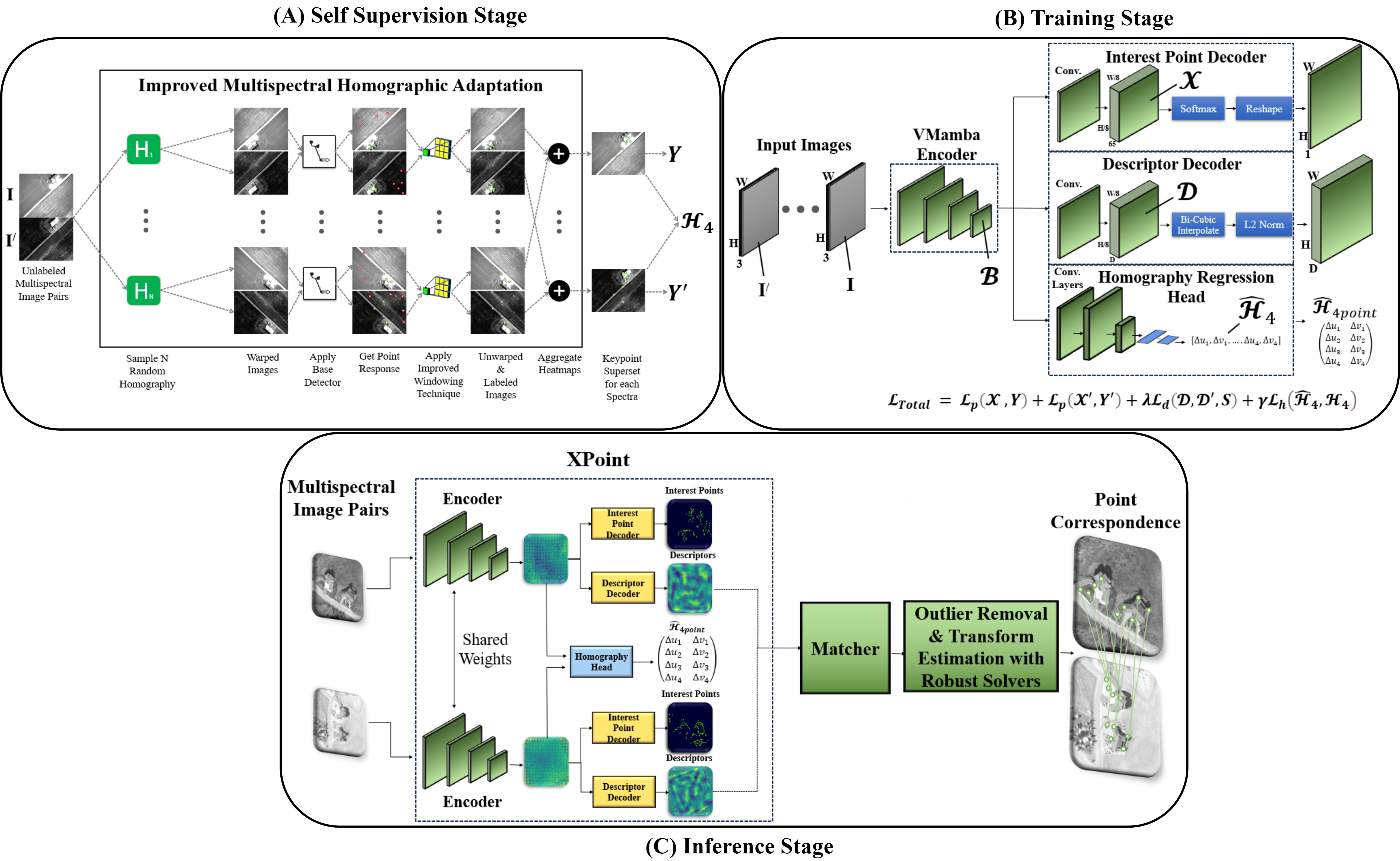}
    \caption{Architecture of XPoint. (A) Self-supervision stage uses improved multispectral homographic adaptation with spectrum-aware keypoint acceptance and the RIFT2 detector to create multispectral pseudo-ground truth keypoints. (B) Training stage combines pretrained VMamba encoders with SS2D for enhanced feature extraction, incorporating interest point and descriptor decoders and a homography regression head for improved matching and homography estimation. (C) In inference stage, shared encoders extract features from multispectral pairs, enabling joint keypoint detection, descriptor extraction, and robust outlier removal for accurate correspondences and homography estimation.}
    \label{fig:overall}
    \vspace{-0.5cm}
\end{figure*}

State-of-the-art image matching methods include hand-crafted and learning-based techniques. Hand-crafted methods like SIFT \cite{SIFT} and SURF \cite{SURF} detect keypoints robust to scale, rotation, and illumination changes but require tuning, are computationally heavy, and struggle with extreme viewpoint and texture variations in multimodal scenarios. The advent of deep learning has led to substantial improvements in addressing these challenges, particularly within the visible spectrum, through learning-based approaches \cite{superpoint,superglue,loftr,aspanformer,gluestick}. Despite these advancements, many learning-based methods encounter difficulties in multimodal settings due to non-linear intensity variations across spectral pairs. To address this, various hand-crafted \cite{multimodal-hc1,multimodal-hc5,multimodal-hc6} and learning-based \cite{multipoint,multimodal-lb2,multimodal-lb3} multispectral image matching frameworks have emerged. While these methods show promise on specific benchmarks, they typically target single-modality scenarios and often struggle to generalize across diverse multimodal tasks. For instance, models like RoMa \cite{roma} that utilize foundational models to enhance generalization still experience issues with overfitting and remain computationally expensive.

This paper introduces XPoint, a modular architecture that enables adaptive training and fine-tuning on new datasets through self-supervision, achieving high performance in multimodal image matching where labeled data is limited. XPoint’s modularity allows rapid adaptability to novel tasks by using aligned image pairs, providing a scalable, self-supervised solution for diverse datasets. Existing methods such as detector-free frameworks \cite{loftr, aspanformer, roma, xoftr} often rely on labeled data (e.g., camera poses, depth maps) unavailable in multimodal contexts, limiting their generalization. XPoint overcomes this by adopting SuperPoint’s self-supervised approach \cite{superpoint} to generate pseudo-ground truth keypoints from aligned pairs, bypassing costly labeling and enhancing cross-modal generalization. Aligned image pairs are considerably easier to obtain than depth maps or precise camera poses, particularly in multimodal settings where additional annotations are scarce or difficult to collect. This self-supervised strategy reduces data collection requirements and enhances generalizability across various modalities. In contrast to similar work, such as ReDFeat \cite{multimodal-lb3}, a self-supervised geometry-informed network that struggles with large viewpoint changes, XPoint effectively manages both viewpoint and spectrum variations.

To handle viewpoint variation and enable self-supervision, SuperPoint \cite{superpoint} uses Homographic Adaptation to create pseudo-ground truth keypoints by simulating different viewpoints of the same scene, allowing a fully convolutional network to recognize scenes from multiple angles. Building on this, MultiPoint \cite{multipoint} adapts Homographic Adaptation for multispectral alignment, combining keypoints from both spectral domains into a unified superset. A key challenge is forming a common feature set across multispectral pairs, which MultiPoint addresses by merging keypoint maps through pixel-wise multiplication and Gaussian smoothing, though some keypoints may be missed. Our previous work, the windowing technique \cite{windowing} refines this approach by allowing cross-spectral keypoints within a defined window around corresponding points in other spectra, allowing small localization errors and enhancing keypoint superset generation in MultiPoint.

This work adopts a self-supervised approach, enhancing pseudo-ground truth keypoint generation with an improved multispectral homographic adaptation stage and introducing a modular architecture with a pre-trained encoder and three joint heads: two dedicated to keypoint detection and description, and a task-specific head—homography regression. The main contributions are summarized as follows (see Fig. \ref{fig:overall} for the complete framework):

\textbf{Multispectral Homographic Adaptation:} A spectrum-aware keypoint acceptance criteria introduces a proposed windowing technique using the RIFT2 multispectral keypoint extractor \cite{rift2}. This component generates viewpoint-invariant pseudo-ground truth keypoint supersets across spectra.  

\textbf{Pretrained VMamba Encoders:} Incorporating 2D-Selective-Scan (SS2D) mechanisms, VMamba encoders provide enhanced semantic awareness, extracting features more effectively than CNNs and more efficiently than visual transformers.

\textbf{Homography Regression Head:} A dedicated task-specific head in the decoder refines keypoint detection and description by guiding homography estimation. This constraint improves matching accuracy, enhances metric performance, and enables flexible model adaptation.

\textbf{Revised Detector Loss:} To address imbalances in challenging datasets like VIS-SAR and VIS-NIR \cite{multimodal-lb3}, a weighted cross-entropy loss is used, boosting confidence and performance under significant spectral differences.

\textbf{Modular XPoint Framework:} Adaptable to various modalities, XPoint achieves state-of-the-art results across four multimodal settings (VIS-NIR, VIS-IR, VIS-LWIR, VIS-SAR) on five datasets \cite{multipoint,vedai,multimodal-lb3}. Using a self-supervised approach with aligned image pairs, XPoint reduces data requirements, enhancing generalization across spectrum shifts and viewpoint changes.

\section{Related Work}
\label{sec:background}

\subsection{Multimodal Image Matching}
Image matching involves tasks like feature matching, patch retrieval, and aligning 2D and 3D point sets within the same spatial frame. In multimodal image matching, these tasks become more challenging due to variations in appearance from different sensors or conditions. Image matching techniques are divided into detector-based and detector-free approaches. Detector-based methods, including pixel-based alignment and feature-based techniques like SIFT \cite{SIFT} and SURF \cite{SURF}, are foundational in fields like medical imaging \cite{R4} but struggle in complex multimodal scenarios. Data-driven methods learn abstract, robust representations, showing superior performance in these settings. Early deep learning approaches, such as \cite{superpoint,d2net,r2d2}, outperformed traditional techniques in multiview tasks. SuperPoint \cite{superpoint} introduced self-supervised training for detectors, generating keypoint heatmaps via Homographic Adaptation and matching with nearest-neighbor and homography solvers. Modern deep learning methods, using graph neural networks, are increasingly replacing traditional pipelines. Techniques like SuperGlue \cite{superglue}, LightGlue \cite{lightglue}, and GlueStick \cite{gluestick} set new standards, with SuperGlue achieving precision at high computational costs, LightGlue reducing costs through adaptive complexity, and GlueStick enhancing matching by integrating interest points and lines.

Detector-free, end-to-end learning methods address the limitations of detector-based approaches, especially in low-texture regions with sparse or ambiguous keypoints. Techniques like \cite{loftr,roma,dkm,pmatch,aspanformer} eliminate explicit keypoint detection by directly learning correspondences, improving performance in challenging settings. LoFTR \cite{loftr} uses a Transformer for coarse-to-fine matching. PMatch \cite{pmatch} enhances textureless region performance through masked modeling and cross-frame transformers. Lightweight models like XFeat \cite{xfeat} improve efficiency, and RoMa \cite{roma} excels in real-world conditions with transformer-based decoding. Despite these advances in visible imagery, multimodal applications remain limited. Hand-crafted methods like POS-GIFT \cite{multimodal-hc1}, RIFT \cite{rift}, and RIFT2 \cite{rift2} bridge this gap by addressing geometric, intensity, and nonlinear radiation distortions (NRD). RIFT2, for example, enhances speed and memory efficiency. Learning-based models \cite{multipoint,multimodal-lb2,multimodal-lb3,xoftr} also outperform traditional techniques in multimodal contexts. For instance, CNN-based models \cite{multimodal-lb2} integrate detection and description in a single forward pass, while ReDFeat \cite{multimodal-lb3} uses a mutual weighting strategy to improve detection and description. XoFTR \cite{xoftr}, based on LoFTR \cite{loftr}, incorporates sub-pixel refinement and scale adjustment in its architecture, enhancing the training process with masked image modeling and pseudo-thermal augmentation.

While detector-free methods like LoFTR \cite{loftr} and PMatch \cite{pmatch} represent a recent trend, they have notable drawbacks. They rely on dense feature matching without keypoint identification, which is computationally intensive, and their generalization is limited by dataset specificity, reducing adaptability to different tasks and environments. Additionally, their reliance on large labeled datasets, which contain information such as depth maps or multiple camera poses, limits their applicability in data-scarce scenarios. To address these challenges, this work adopts a self-supervised detector-based approach inspired by SuperPoint \cite{superpoint}, utilizing homographic adaptation to generate task-specific pseudo-ground truth keypoints, enhancing adaptability across modalities and simulating various viewpoints. While SuperPoint advanced multi-view matching, it faced limitations in multimodal scenarios. MultiPoint extends SuperPoint’s framework to multispectral images, applying homographic adaptation to optical-thermal pairs by merging keypoint maps generated with the SURF detector and Gaussian filtering to improve cross-modal matching. However, the convolutional encoders in SuperPoint and MultiPoint struggle with long-range dependencies, limiting their effectiveness in capturing broader contextual information for robust keypoint detection. This paper describes our solution to these problems.

\vspace{-0.3cm}
\subsection{Homography}
A homography is a \(3 \times 3\) matrix, \(\mathcal{H}\), that transforms points and lines between two image planes, preserving straight lines. It is essential in tasks like image stitching and perspective correction. Although \(\mathcal{H}\) has nine elements, it possesses eight degrees of freedom due to scaling. Intuitively, \(\mathcal{H}\) represents translation, rotation, scaling, and perspective distortion. The relationship between pixel coordinates in two images is given by:

\vspace{-0.2cm}
\begin{equation}\label{hm-matrix}
\begin{pmatrix}
u' \\
v' \\
1
\end{pmatrix}
=
\begin{pmatrix}
\mathcal{H}_{11} & \mathcal{H}_{12} & \mathcal{H}_{13} \\
\mathcal{H}_{21} & \mathcal{H}_{22} & \mathcal{H}_{23} \\
\mathcal{H}_{31} & \mathcal{H}_{32} & \mathcal{H}_{33}
\end{pmatrix}
\begin{pmatrix}
u \\
v \\
1
\end{pmatrix}
\end{equation}
where \((u, v)\) are original pixel coordinates, and \((u', v')\) are transformed coordinates. The submatrix \([\mathcal{H}_{11}, \mathcal{H}_{12}; \mathcal{H}_{21}, \mathcal{H}_{22}]\) represents rotation and scaling, \([\mathcal{H}_{13}, \mathcal{H}_{23}]\) translation, and \([\mathcal{H}_{31}, \mathcal{H}_{32}]\) perspective distortion.

\subsubsection{Homographic Adaptation}
Homographic Adaptation \cite{superpoint} enhances viewpoint-invariant interest point detection by applying random homographies to an input image, extracting and combining keypoints to form a superset, thereby improving detection accuracy under geometric distortions. Let \( f \) denote the base keypoint detector, \( I \) the input image, and \( x = f(I) \) the interest points. For a random homography \( \mathcal{H}_i \), ideally, the detector is covariant: \( \mathcal{H}x = f(\mathcal{H}(I)) \) or equivalently \( x = \mathcal{H}^{-1} f(\mathcal{H}(I)) \). Since detectors are not fully covariant, Homographic Adaptation mitigates this by averaging over multiple random homographies, yielding an improved detector:
\begin{equation}
\hat{F}(I; f) = \frac{1}{N_{h}} \sum_{i=1}^{N_{h}} \mathcal{H}_{i}^{-1} f(\mathcal{H}_{i}(I))
\end{equation}

\subsubsection{Multispectral Homographic Adaptation} 
Multispectral Homographic Adaptation \cite{multipoint} extends homographic adaptation to multimodal images, detecting points invariant to both perspective changes and spectrum differences. Random homographies are applied to multispectral pairs to generate interest point heatmaps, which are aggregated using pixel-wise multiplication to ensure consistency across spectra and perspectives. The final function \( \hat{F} \) aggregates heatmaps from optical and thermal pairs, producing cross-spectral interest points:
\vspace{-0.2cm}
\begin{equation}
\hat{F}(I_o, I_t, f) = \frac{1}{N_h} \sum_{i=1}^{N_h} \mathcal{H}_i^{-1} \left( f \left( \mathcal{H}_i \left( I_o \right) \right) \odot f \left( \mathcal{H}_i \left( I_t \right) \right) \right),
\end{equation}
where \( I_o \) and \( I_t \) denote the optical and thermal images, respectively. Each homography \( \mathcal{H}_i \) simulates a unique viewpoint transformation and, when applied to \( I_o \) and \( I_t \), generates transformed versions of these images. The detector outputs from these transformed images undergo an operation \( \odot \), which represents element-wise multiplication in this case, after which the resulting features are mapped back to the original image space using the inverse homography \( \mathcal{H}_i^{-1} \). Finally, averaging over all \( N_h \) outputs produces \( \hat{F}(I_o, I_t, f) \), providing a robust feature matching function between the optical and thermal images. MultiPoint \cite{multipoint} achieves cross-spectral alignment by overlapping keypoints through pixel-wise multiplication, though some keypoints may be missed. To address this, \cite{windowing} introduces a windowing technique that accepts cross-spectral keypoints within a defined window around corresponding locations, improving homography estimation and generating enhanced keypoint sets for training.

\subsubsection{Homography Estimation}
Homography estimation is central in computer vision tasks such as camera calibration, image registration, pose estimation, and SLAM. It maps two images of the same planar surface from different perspectives, revealing camera pose transformations. Traditional methods include feature-based approaches (e.g., SIFT, SURF with RANSAC) and direct methods (e.g., Lucas-Kanade for pixel-to-pixel matching). Recently, deep learning methods like HomographyNet \cite{detone} directly learn mappings between image pairs and homographies, bypassing traditional feature extraction. Unsupervised and self-supervised models like SSR-Net \cite{invertibility} minimize pixel-wise errors to reduce label dependence. Other advancements include hierarchical models \cite{hierarchical}, attention-based methods \cite{attention-based-hm}, and iterative multi-scale models \cite{iterative,ms2ca} for improved robustness in complex scenes. For multimodal homography estimation, VisIRNet \cite{VisIRNetSedatÖzer} surpasses traditional methods by directly estimating the homography matrix and image corners, enhancing performance on multimodal datasets.

\vspace{-0.2cm}

\section{XPoint Architecture}
We introduce XPoint, a self-supervised framework developed to address multimodal image-matching challenges through a modular architecture. The framework operates in three stages: multispectral homographic adaptation, training, and inference (see Fig. \ref{fig:overall}). First, pseudo-ground truth keypoints are generated using the proposed multispectral homographic adaptation (see Fig. \ref{fig:overall}(A)). This process applies spectrum-aware keypoint acceptance criteria and utilizes a base detector like RIFT2 \cite{rift2} to create pseudo-ground truth keypoint supersets that are invariant to both viewpoint and spectrum variations across different modalities.

In the training stage (see Fig. \ref{fig:overall}(B)), the network processes multispectral image pairs to generate keypoint heatmaps, descriptors, and homography parameters. The core VMamba encoder \cite{vmamba}, pre-trained on segmentation tasks, enhances the network’s capability to align multispectral features in a shared space, improving modality alignment. Compared to traditional CNNs or visual transformers like SwinV2 \cite{swinv2}, this architecture optimally balances efficiency and accuracy. Additionally, a lightweight homography regression head guides keypoint detection and description, ensuring geometric consistency and boosting matching accuracy. This modular head is adaptable, supporting flexible configuration. In datasets like VIS-SAR, with significant nonlinear radiation distortions (NRD), a weighted cross-entropy loss manages false positives, enabling the model to prioritize learning keypoints that strengthen feature matching even under challenging conditions.

In the inference stage (see Fig. \ref{fig:overall}(C)), the detected keypoints can be matched using traditional methods like Mutual Nearest Neighbor with RANSAC or MAGSAC, or deep learning-based matchers such as LightGlue \cite{lightglue}. This modular design offers flexibility, enabling users to adjust or replace components to suit specific tasks or datasets. Overall, XPoint’s three-stage, modular architecture effectively addresses the complexities of multimodal image matching, providing adaptability for applications ranging from feature matching to homography estimation across diverse spectral modalities.

\vspace{-0.2cm}
\begin{figure}[h]
    \centering
    \includegraphics[width=\linewidth]{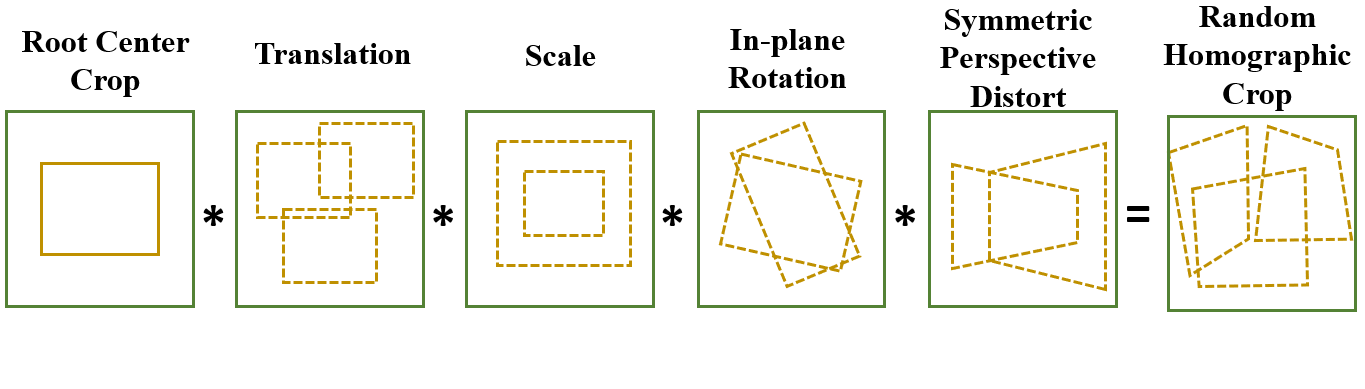}
    \vspace{-1cm}
    \caption{Random homographies are generated by combining simple transformations such as translation, scaling, rotation, and symmetric perspective distortion, sampled within predefined ranges.}
    \label{fig:choose_homography}
    \vspace{-0.6cm}
\end{figure}

\subsection{Improved Multispectral Homographic Adaptation}
In this work, an enhanced multispectral homographic adaptation technique (see Fig. \ref{fig:overall}(A)) is employed to generate ground truth keypoints for multispectral image pair datasets. The foundation of this method is the ``windowing'' technique \cite{windowing}, originally developed to refine keypoint acceptance criteria across different spectral channels. Unlike simpler methods that directly sum or multiply probability heatmaps for each homography-warped pair, the windowing technique identifies corresponding keypoints in another spectrum within a predefined window around a specific keypoint location. In the original implementation \cite{windowing}, keypoints were accepted if they met the window criteria, with an assigned probability of ``1'' to these points. In our approach, we enhance this by treating these values as probabilities that accumulate over multiple observations. Rather than binary acceptance, we aggregate keypoint probabilities across spectra and filter them based on a threshold at the end. This adjustment ensures that keypoints not only meet window criteria across spectra but also remain consistently detected across multiple viewpoints, increasing robustness to viewpoint variations. The resulting pseudo-ground truth keypoints are thus resilient to spectral changes, with adjustable window sizes, and to viewpoint variations, enhancing self-supervised learning performance.

We adopt the strategy outlined by the authors of SuperPoint \cite{superpoint} for selecting appropriate homographies in Homographic Adaptation. Since not all 3x3 matrices represent plausible camera transformations, we decompose the homography into interpretable transformation classes, including translation, scaling, in-plane rotation, and symmetric perspective distortion. Each transformation is sampled within predefined ranges using a truncated normal distribution. These transformations are then combined with an initial root center crop to reduce bordering artifacts, as illustrated in Figure \ref{fig:choose_homography}.

\begin{figure}[h]
    \centering
    \includegraphics[width=\linewidth]{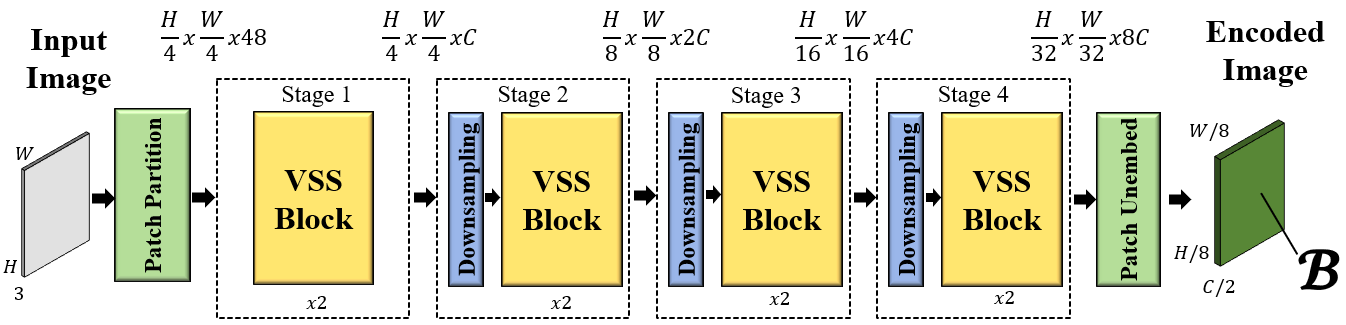}
    \caption{Adopted VMamba Encoder Architecture.}
    \label{fig:adopted-vmamba}
    \vspace{-0.5cm}
\end{figure}

\subsection{VMamba as an Encoder}
To enhance feature extraction for multispectral image matching, we integrate the VMamba encoder, a state-of-the-art network optimized for computational efficiency and high accuracy with multimodal data. Our modular architecture allows the encoder to be easily replaced or adapted as needed. For this work, we select the VMamba-T[s1l8] \cite{vmamba} variant, pretrained on ADE20K for semantic segmentation, enabling it to capture semantically meaningful features across modalities and reducing the domain gap between spectral bands. This choice balances accuracy and computational efficiency, offering a robust alternative to traditional CNNs and visual transformers. A key feature of VMamba is its 2D-Selective-Scan (SS2D) mechanism, which focuses computational resources on the most relevant image regions, optimizing feature extraction and reducing the overhead associated with high-resolution images. This makes VMamba particularly suited for aligning and matching high-dimensional multispectral image pairs with enhanced precision and efficiency.

The VMamba encoder processes input images through a four-stage design that progressively refines feature representations across scales (see Fig. \ref{fig:adopted-vmamba}). The encoder receives an image of dimensions \( H \times W \times 3 \), representing RGB or multispectral inputs. First, the input image is divided into patches, creating an initial tensor of \( \frac{H}{4} \times \frac{W}{4} \times 48 \), reducing spatial resolution but preserving essential local information. Successive stages, referred to as Stage 1 through 4, apply downsampling and Visual State Space (VSS) blocks to refine features at progressively smaller scales and higher channel depths, ultimately yielding a compact multi-scale feature map of size \( \frac{H}{8} \times \frac{W}{8} \times \frac{C}{2} \), suitable for downstream tasks.

The modular design of our architecture permits encoder substitution to meet task-specific needs. While VMamba is used here for its blend of semantic feature extraction and efficiency, alternative backbones, like Swin Transformers or CNNs, can be integrated as required. VMamba's selection is driven by its semantic awareness, efficiency with SS2D, and balanced performance, offering a strong alternative between CNNs and visual transformers. The VMamba encoder output, denoted \( \mathcal{B} \in \mathbb{R}^{\frac{H}{8} \times \frac{W}{8} \times \frac{C}{2}} \), provides a dense, multi-scale feature representation that captures spatial and semantic information, facilitating keypoint detection and matching tasks.

\begin{figure}[h]
    \centering
    \includegraphics[width=\linewidth]{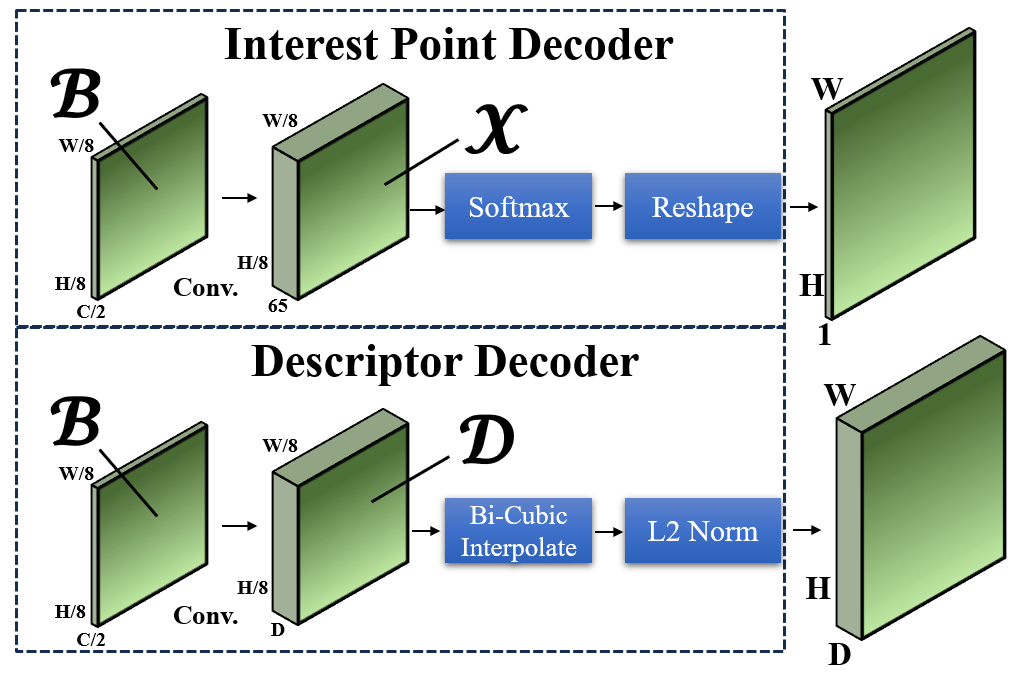}
    \caption{Interest Point and Descriptor Decoders. The interest point decoder outputs keypoint heatmaps, while the descriptor decoder generates dense descriptors for the input image.}
    \label{fig:decoders}
    \vspace{-0.5cm}
\end{figure}

\subsection{Interest Point and Descriptor Decoders}
\label{decoderstage}
\noindent\textbf{Interest Point Decoder.} 
The interest point decoder takes the intermediate tensor \( \mathcal{B} \in \mathbb{R}^{\frac{H}{8} \times \frac{W}{8} \times \frac{C}{2}} \) from the encoder and applies a convolutional layer to compute the tensor \( \mathcal{X} \in \mathbb{R}^{\frac{H}{8} \times \frac{W}{8} \times 65} \). The 65 channels in this tensor correspond to different non-overlapping $8 \times 8$ pixel blocks, with each of the first 64 channels representing potential interest point locations in one of the blocks, and the 65th channel, ``dustbin'' cell, indicating blocks where no keypoints are present. After computing \( \mathcal{X} \), a softmax function is applied along the channel dimension, transforming the raw values into probabilities. This process ensures that the network treats keypoint detection as a classification task, where each pixel block is classified into one of 65 classes. The first 64 channels represent the presence of keypoints in their corresponding locations within $8 \times 8$ pixel blocks, while the 65th channel indicates the absence of keypoints. The resulting tensor is reshaped to match the original image size \( H \times W \) by selecting the most probable keypoint class/location per block and discarding the ``dustbin'' channel. The final output is a heatmap of keypoint locations across the entire image. 

% This single-label, multi-class classification-based design allows the model to predict multiple keypoints within the same block, leveraging soft assignment to distribute probabilities across candidate keypoints. This strategy provides a more flexible and informative representation than the hard assignment proposed in SuperPoint\cite{superpoint} and Multipoint\cite{multipoint}, which limits predictions to a single keypoint per block, which is a single-label multi-class classification. The final output is a heatmap of keypoint locations across the entire image.

\noindent\textbf{Descriptor Decoder.} 
The descriptor decoder similarly takes the intermediate tensor \( \mathcal{B} \) from the encoder and applies a convolutional layer to compute the tensor \( \mathcal{D} \in \mathbb{R}^{\frac{H}{8} \times \frac{W}{8} \times D} \), where \( D \) is the fixed length of the descriptor. This tensor provides a semi-dense grid of descriptors, with each descriptor associated with a specific $8 \times 8$ pixel region in the original image. To produce dense descriptors, the output of the descriptor decoder undergoes bicubic interpolation, which upscales the semi-dense grid to match the original image size \( H \times W \). Following the interpolation, L2 normalization is applied to each descriptor, ensuring that all descriptors lie on the unit hypersphere in \( \mathbb{R}^{D} \). This normalization step is essential for stabilizing the matching process, as it ensures that descriptor comparisons are based solely on the angular similarity between vectors. The use of semi-dense descriptors strikes a balance between memory efficiency and performance. Fully dense descriptor extraction can be computationally expensive and memory-intensive, particularly for high-resolution images. By focusing on one descriptor per $8 \times 8$ pixel block, the decoder reduces the computational burden while maintaining sufficient spatial resolution for effective matching.

% \subsection{Homography Regression Head}
% In this paper, we employ a joint keypoint extraction and homography estimation framework, as opposed to the typical two-stage approach where keypoints are detected first. Then homography is estimated by matching selected keypoints. We observe that instead of computing homography using feature matching techniques, directly estimating the homography using deep regression network could be achieved by  multitask learning, which is applied to jointly train models for homography regression, keypoint detection and description generation. By sharing representations and parameters across the multiple tasks, the model can learn to detect keypoints and describe their local features while simultaneously estimating the homography transformation between the images. This leads to improved performance on all tasks, as the model learns to extract more informative and robust features from the images that are relevant to all tasks.

\vspace{-0.2cm}
\begin{figure}[h]
    \centering
    \includegraphics[width=\linewidth]{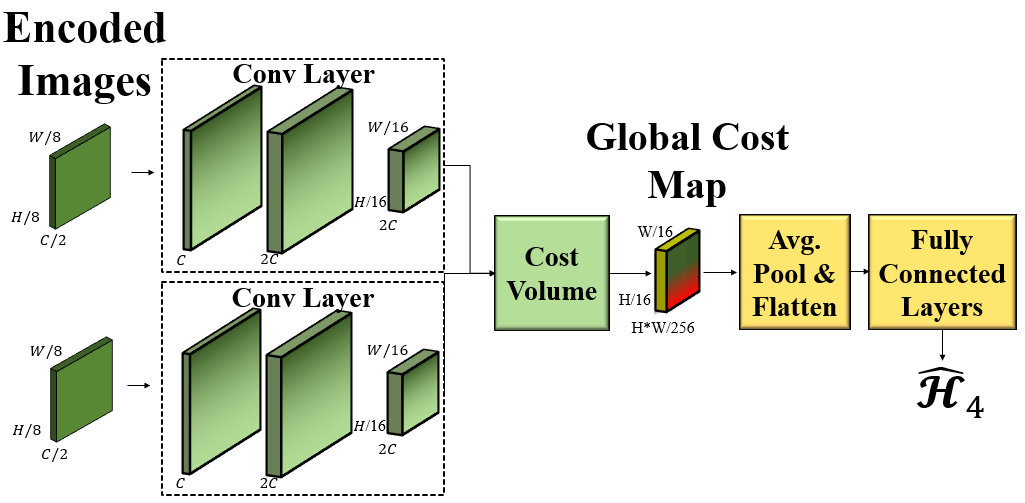}

    \caption{Homography Head Architecture.}
    \label{fig:hmhead}
    \vspace{-0.5cm}
\end{figure}

\subsection{Homography Regression Head}
In addition to the interest point and descriptor heads, we introduce a task-specific homography regression head to our design. Rather than directly estimating the final homography, this head imposes a homography constraint during training. This constraint enables multitask learning, allowing for the joint training of keypoint detection, descriptor generation, and homography constraint application. By sharing representations and parameters across these tasks, the model learns more robust and informative features, improving performance in keypoint extraction and description while maintaining geometric consistency between images. Additionally, multitask learning \cite{multitask} significantly reduces overfitting in hard parameter-sharing models like ours (see Fig. \ref{fig:overall}). The homography regression head builds on methodologies from DIHE \cite{detone} and SRHEN \cite{srhen}, following DIHE's supervision and training approach and incorporating an attention-based cost volume inspired by SRHEN. Further details on these aspects are elaborated in the training methodology section.

Our homography regression head (see Fig. \ref{fig:hmhead}) processes two input image streams through a neural network. The first stage, Layer1, applies convolutional layers with $3 \times 3$ kernels, batch normalization, and ReLU activation, followed by max pooling to produce feature maps. These feature maps are used to compute a cost volume via batch matrix multiplication, capturing feature similarities between the images. This cost volume aids in homography estimation and is condensed with adaptive average pooling before being passed through fully connected layers with dropout for regularization. The output is a flattened version of the 4-point parameterization matrix, $\mathcal{H}_4$, from which the homography matrix $\mathcal{H}$ is computed (see Section \ref{fourpoint}).

% Mathematical description of the cost volume, akin to attention mechanism
Given feature maps $x_1$ and $x_2$ from two input images, the cost volume $CV$ is computed by taking the dot product of feature vectors along the depth dimension:
\begin{equation}
    CV(i, j) = \sum_{k} \frac{x_1(i, k) \cdot x_2(j, k)}{\|x_1(i, k)\| \|x_2(j, k)\|},
\end{equation}
where $i$ and $j$ denote spatial positions within the feature maps, and $k$ traverses the feature channels. This formula assesses feature similarity at various spatial locations, similar to attention scores, aiding in accurate homography estimation between the two images.

\vspace{-0.2cm}
\section{Experimental Setup}
\label{experimental_setup}
The experimental evaluation uses multiple multimodal datasets. The primary dataset, the Multispectral (Optical-Thermal) Image Pair Dataset by Multipoint \cite{multipoint}, includes 9340 training and 4391 testing image pairs focused on aerial agricultural data. The VEDAI dataset \cite{vedai} contains 1245 NIR (0.75–1.4 $\mu$m) image pairs, with 1045 for training and 200 for testing, used for urban area detection in aerial imagery. Additional datasets from ReDFeat \cite{multimodal-lb3} include VIS-NIR, VIS-IR, and VIS-SAR datasets. VIS-NIR contains 345 training and 128 testing pairs from scenes like countryside and urban areas, VIS-IR includes 221 training and 47 testing pairs of visible and infrared images, and VIS-SAR has 2011 training and 424 testing pairs of visible and SAR images. These datasets allow comprehensive evaluation of model performance in real-world multimodal scenarios. The MultiPoint training set, the largest available, is used for base training to establish a baseline model, which is subsequently fine-tuned using the other training sets.

% Summary of experimental setup can be seen in Figure \ref{fig:experimental_setup}

\vspace{-0.3cm}
\subsection{Ground Truth Data Extraction}
\textbf{Keypoint selection.} RIFT2 \cite{rift2} is selected as the base detector due to its significant improvements over traditional methods like SIFT and SURF, particularly in handling nonlinear radiation distortions (NRD) common in multispectral image matching. Unlike SIFT\cite{SIFT}, which is sensitive to these distortions, and SURF\cite{SURF}, which may struggle with rotation and scale variances, RIFT2 enhances the robustness of the original RIFT \cite{rift} by incorporating a more efficient rotation-invariance technique. Training datasets are labeled using RIFT2 combined with the proposed homographic adaptation technique. The window size for the proposed windowing technique is set to 5, and 100 random homographies are applied in the multispectral homographic adaptation. Random homographies are generated by decomposing them into simpler transformations: translation (\(\pm \%5\)), scale (\(\pm \%20\)), rotation (\(\pm 90^\circ\)), and distortion (\(\pm \%20\) in width and height), resulting in complex transformations. Border reflection is used during training and testing to avoid artificial boundaries in warped images.

\noindent \textbf{4-Point Homography Parameterization.}
\label{fourpoint}
In image alignment, homographies are typically represented by a 3x3 matrix, \(\mathcal{H}\), transforming pixel coordinates from \([u, v]\) in one image to \([u', v']\) in another (see Equation \ref{hm-matrix}). This standard approach, which includes rotational, translational, and perspective elements, can complicate the optimization process.

To simplify, an alternate parameterization focused on corner locations, known as the 4-point parameterization \cite{detone}, is adopted. This approach reduces the representation to corner displacements \((\Delta u_1, \Delta v_1, \ldots, \Delta u_4, \Delta v_4)\) as \(\mathcal{H}_{4point}\), expressed as:

\vspace{-0.2cm}
\begin{equation}\label{h4points}
\mathcal{H}_{4point} = 
\begin{pmatrix}
\Delta u_1 & \Delta v_1 \\
\Delta u_2 & \Delta v_2 \\
\Delta u_3 & \Delta v_3 \\
\Delta u_4 & \Delta v_4
\end{pmatrix}
\end{equation}
where \(\Delta u_i = u'_i - u_i\) and \(\Delta v_i = v'_i - v_i\) represent horizontal and vertical displacements for each corner point. This 4-point method simplifies the standard \(3 \times 3\) homography matrix by focusing on corner displacements, making homography more intuitive for alignment. Conversion from \(\mathcal{H}_{4point}\) to \(\mathcal{H}_{matrix}\) can be achieved using the normalized Direct Linear Transform (DLT), which is a well-established algorithm for estimating projective transformations \cite{hartley2003multiple}.

\vspace{-0.3cm}
\subsection{Self-Supervision and Training}
\textbf{Self-Supervision.} 
During training, image pairs linked by a known homography and labeled with keypoint locations from RIFT2 \cite{rift2} are utilized. Pre-aligned image pairs from the same or different spectra are randomly selected, with one image warped by a random known homography, in order to provide balanced exposure to multispectral and same-spectrum pairs during training. The model is trained with PyTorch, using the Adam optimizer, for 3000 epochs on the Multispectral (optical-thermal) Image Pair Dataset, with a batch size of 32, automatic mixed precision, and a learning rate of 0.001. For other datasets, the base model is fine-tuned for 100 epochs. Training incorporates translation, rotation, scaling, and perspective changes for homography sampling to enhance model robustness. Additionally, $256\times256$ patches are randomly cropped from full-resolution images ($512\times640$) as data augmentation.

\noindent\textbf{Encoder and Decoder Training.} 
The encoder, fine-tuned from a pretrained VMamba state, and the decoder, trained from scratch, are simultaneously trained to optimize the encoder’s advanced features for multispectral matching.

\noindent\textbf{Homography Regression Head Training.} 
Ground truth corner displacements are derived by applying the ground truth homography transformation to the first image’s corners, as described in Equation \ref{hm-matrix}. Using the encoded image pairs as input, the homography regression head computes prediction of these eight displacement values. The homography head is optimized with Euclidean (L2) loss between ground truth and predicted displacements during training.

\vspace{-0.3cm}
\subsection{Loss Functions}
The overall loss function, $\mathcal{L}$, is a weighted sum of three components: $\mathcal{L}_{p}$ for the interest point detector, $\mathcal{L}_{d}$ for the descriptor, and $\mathcal{L}_{h}$ for homography regression. Given pairs of multispectral images related by randomly generated homographies, we obtain pseudo-ground truth interest point locations and correspondences via homography $\mathcal{H}$. The loss function is defined as:

\vspace{-0.4cm}
\begin{equation}
\mathcal{L} = \mathcal{L}_{p}(\mathcal{X}, Y) + \mathcal{L}_{p}(\mathcal{X}^{\prime}, Y^{\prime}) + \lambda \mathcal{L}_{d}(\mathcal{D}, \mathcal{D}^{\prime}, S) + \gamma \mathcal{L}_{h}(\hat{\mathcal{H}}_4, \mathcal{H}_4)
\end{equation}
where \(\mathcal{X}\) and \(\mathcal{X}^{\prime}\) represent the keypoint tensors in the original and transformed images, respectively, while \(\mathcal{D}\) and \(\mathcal{D}^{\prime}\) are the descriptor tensors for these images (see Section \ref{decoderstage}). \(Y\) and \(Y^{\prime}\) denote the ground truth interest point locations in the original and transformed images, respectively, and \(S\) denotes the entire set of correspondences for a pair of images. \(\mathcal{H}_4\) is the ground truth 4-point homography parameterization, with \(\hat{\mathcal{H}}_4\) as the predicted homography.

\subsubsection{Interest Point Detector Loss}
The loss function for the interest point detector, \( \mathcal{L}_{p} \), is defined as a single-label, multi-class cross-entropy loss with adjustable weights to address class imbalance, especially in challenging datasets like VIS-SAR where keypoints may be sparse or absent in one spectrum. The 65th channel, known as the dustbin class, handles cases without keypoints, and its weight \( w_{65} \) is a hyperparameter optimized for challenging spectral conditions. This adjustable weight allows fine-tuning of the model’s sensitivity to false positives, encouraging accurate keypoint predictions even in regions without ground truth markers.

The interest point detection loss \( \mathcal{L}_{p} \) is computed over cells \( \mathbf{x}_{hw} \) within the predicted tensors \( \mathcal{X} \) and \( \mathcal{X}' \), each sized \( \frac{H}{8} \times \frac{W}{8} \times 65 \), for two input images with ground truth tensors \( Y \) and \( Y' \):
\vspace{-0.2cm}
\begin{equation}
\mathcal{L}_{p}(\mathcal{X}, Y) = \frac{1}{H_c W_c} \sum_{h=1}^{H_c} \sum_{w=1}^{W_c} l_p\left(\mathbf{x}_{hw}, y_{hw}\right)
\end{equation}

where
\vspace{-0.2cm}
\begin{equation}
l_p\left(\mathbf{x}_{hw}, y_{hw}\right) = - \sum_{c=1}^{65} w_c \, y_{hwc} \log \left( \frac{\exp(x_{hwc})}{\sum_{i=1}^{65} \exp(x_{hwi})} \right).
\end{equation}

Here, \( \mathbf{x}_{hw} \) represents predicted logits for cell \( (h, w) \), and \( y_{hw} \) provides the ground truth one-hot encoded locations. The weight \( w_c \) modulates the contribution of each class, particularly \( w_{65} \) for the dustbin class, typically set at $0.025$ in the VIS-SAR setting.

\noindent\textbf{Keypoint Cell Computation.}  
The ground truth tensor \( Y \) is derived by dividing the input image into non-overlapping \( 8 \times 8 \) pixel regions. Each block is represented by a one-hot encoded vector across 65 channels: the first 64 for keypoints, and the 65th for the dustbin cell, indicating no keypoints. If multiple keypoints appear within a block, one is randomly selected.

\subsubsection{Descriptor Loss}
The descriptor loss compares pairs of descriptor cells, with $\mathbf{d}_{h w}$ in $\mathcal{D}$ from the first image and $\mathbf{d}^{\prime}_{h^{\prime}w^{\prime}}$ in $\mathcal{D}^{\prime}$ from the second spectrum. Correspondence between a cell at $(h, w)$ in the first spectrum and its matching cell at $(h^{\prime}, w^{\prime})$ in the second spectrum is defined by the homography as follows:
\vspace{-0.2cm}
\begin{equation}
s_{h w h^{\prime} w^{\prime}}= \begin{cases}1, & \text { if }\left\|\widehat{\mathcal{H} \mathbf{p}_{h w}}-\mathbf{p}_{h^{\prime} w^{\prime}}\right\| \leq 4 \\ 0, & \text { otherwise }\end{cases}
\end{equation}
where $\mathbf{p}_{hw}$ is the central pixel in cell $(h, w)$, and $\widehat{\mathcal{H} \mathbf{p}_{h w}}$ is the transformed position under homography $\mathcal{H}$. A weighting factor $\lambda_{d}$ balances the descriptor loss, with hinge loss margins $m_{p}$ and $m_{n}$ applied to positive and negative matches, respectively:
\vspace{-0.1cm}
\begin{equation}
\begin{aligned}
& \mathcal{L}_{d}\left(\mathcal{D}, \mathcal{D}^{\prime}, S\right)= \\
& \frac{1}{\left(H_{c} W_{c}\right)^{2}} \sum_{\substack{h=1 \\
w=1}}^{H_{c}, W_{c}} \sum_{\substack{h^{\prime}=1 \\
w^{\prime}=1}}^{H_{c}, W_{c}} l_{d}\left(\mathbf{d}_{h w}, \mathbf{d}_{h^{\prime} w^{\prime}}^{\prime} ; s_{h w h^{\prime} w^{\prime}}\right),
\end{aligned}
\end{equation}
where
\vspace{-0.1cm}
\begin{equation}
\begin{array}{r}
l_{d}\left(\mathbf{d}, \mathbf{d}^{\prime} ; s\right)=\lambda_{d} * s * \max \left(0, m_{p}-\mathbf{d}^{T} \mathbf{d}^{\prime}\right) \\
+(1-s) * \max \left(0, \mathbf{d}^{T} \mathbf{d}^{\prime}-m_{n}\right).
\end{array}
\end{equation}

\subsubsection{Homography Regression Head Loss}
The homography regression head loss, $\mathcal{L}_{h}$, uses Euclidean (L2) loss to compare the ground truth 4-point parametrization $\mathcal{H}_4$ of the homography matrix $\mathcal{H}$ and the predicted 4-point parametrization $\hat{\mathcal{H}_4}$. The loss is given by:
\vspace{-0.1cm}
\begin{equation}
\mathcal{L}_{h}(\hat{\mathcal{H}}_4,\mathcal{H}_4) = \sum_{i=1}^{4} {||\mathcal{H}_{4}^i - \hat{\mathcal{H}}_{4}^i||}^2_2
\end{equation}
where $\mathcal{H}_{4}^i$ and  $\hat{\mathcal{H}}_{4}^i$ are the $i^{th}$ corner points in the ground truth and predicted 4-point parametrization, respectively.

\vspace{-0.2cm}
\subsection{Evaluation Protocol}
\label{evalmetrics}

In the evaluation phase, the model is compared against both detector-based and detector-free methods using the protocol by Mikolajczyk et al. \cite{eval}. This evaluation involves computing keypoints and their descriptors, with random homographies applied to generate transformations using translation (\(\pm 5\%\)), scale (\(\pm 10\%\)), rotation (\(\pm 90^\circ\)), and symmetric distortion (\(\pm 5\%\) in width and height). These transformations create warped test sets for performance assessment.

For detector-based methods, keypoints and descriptors are first extracted and then matched using a bidirectional nearest neighbor strategy. Detector-free methods directly predict matches. Metrics such as Repeatability and Matching Score are used to evaluate feature matching accuracy, with MAGSAC \cite{magsac} applied in both approaches to remove outliers and refine matches, enabling accurate homography estimation. A MAGSAC reprojection threshold of 2 is used for all experiments. The following evaluation metrics are used in this study (see \cite{multipoint} for details):

\textbf{Repeatability:} It measures the model’s ability to detect the same interest points in both images. It is calculated based on the consistency of keypoints within overlapping regions, with a tolerance threshold of 5 pixels.

\textbf{Matching Score:} It evaluates the ratio of correctly matched keypoints to the total detected features within the overlapping view area. This metric is calculated symmetrically for both images and then averaged, using a 5-pixel threshold to assess matching accuracy.

% \textbf{Homography Estimation:} Homography Estimation assesses alignment accuracy by measuring the deviation of transformed image corners from their expected positions according to the ground truth homography. A variable pixel distance threshold, ranging from 1 to 10 pixels, is used to evaluate the precision of the alignment.

\textbf{Homography Estimation:} It measures image alignment accuracy by calculating the fraction of transformed corners within a pixel threshold $\varepsilon$ of their ground truth positions:

\begin{equation}\label{homoest}
\frac{1}{N} \sum_{i=1}^{N} \mathbb{1} \left( dist(corner_i, corner_{i,gt}) \leq \varepsilon \right)
\end{equation}

Here, \textit{$\text{corner}_i$} and \textit{$\text{corner}_{i,\text{gt}}$} are the transformed and ground truth corners, respectively, \textit{dist} is the Euclidean distance, $\mathbb{1}$ is the indicator function, and $N$ is the number of corners (typically 4). The threshold $\varepsilon$ ranges from 1 to 10 pixels to evaluate alignment precision.

\vspace{-0.2cm}
\section{Experiments}
\begin{figure*}[t]
    \centering
    \includegraphics[width=1\textwidth]{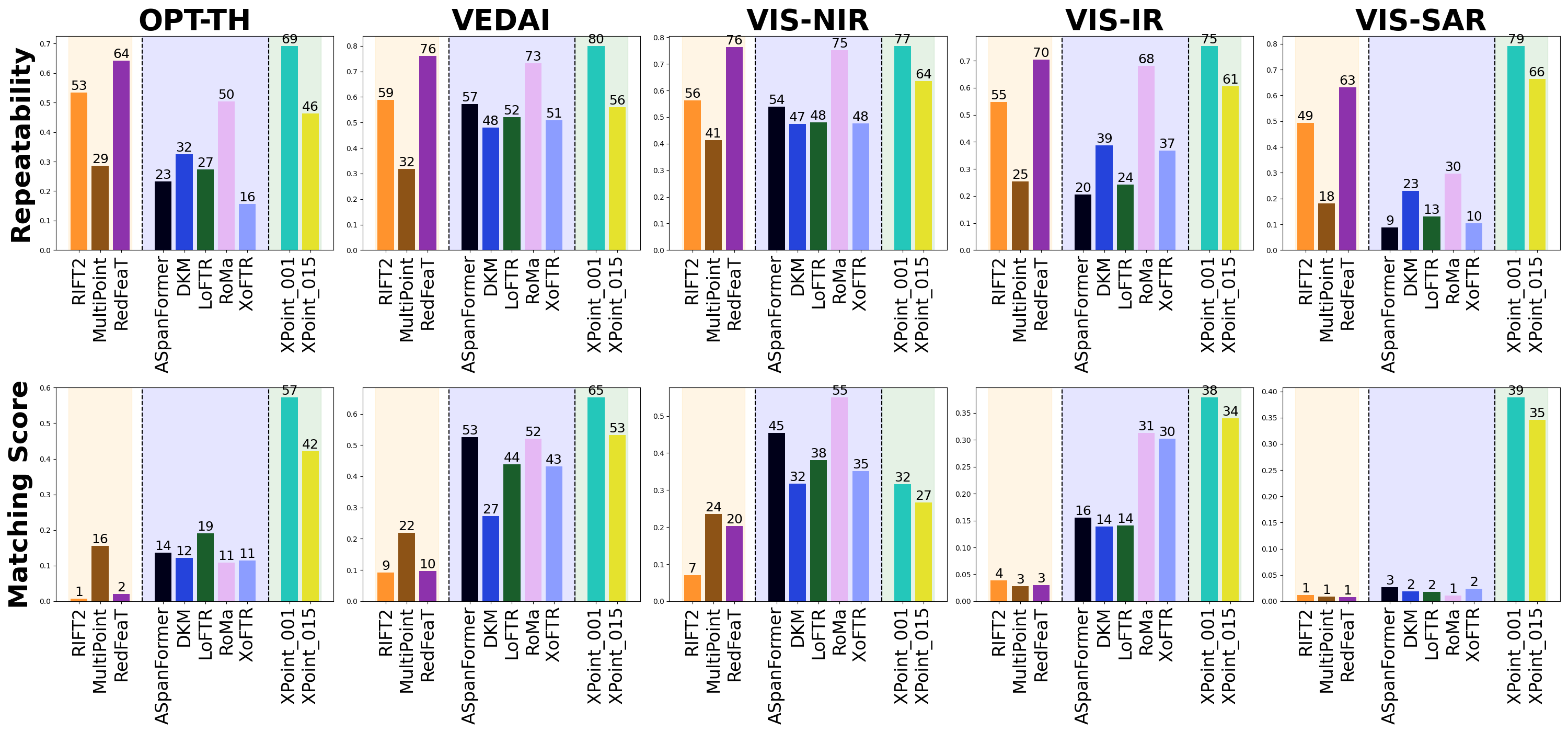}
\caption{Feature Matching Performance evaluated with Repeatability and Matching Score across datasets. XPoint variants demonstrate superior repeatable keypoint detection and matching consistency, with XPoint\_001 outperforming state-of-the-art models like RoMa and ReDFeat on most datasets, except VIS-NIR, where detector-free methods benefit from similar training and test domains.}
    \label{fig:rr_and_mscore}
    \vspace{-0.5cm}
\end{figure*}

\begin{figure}[h]
    \centering
    \includegraphics[width=\columnwidth]{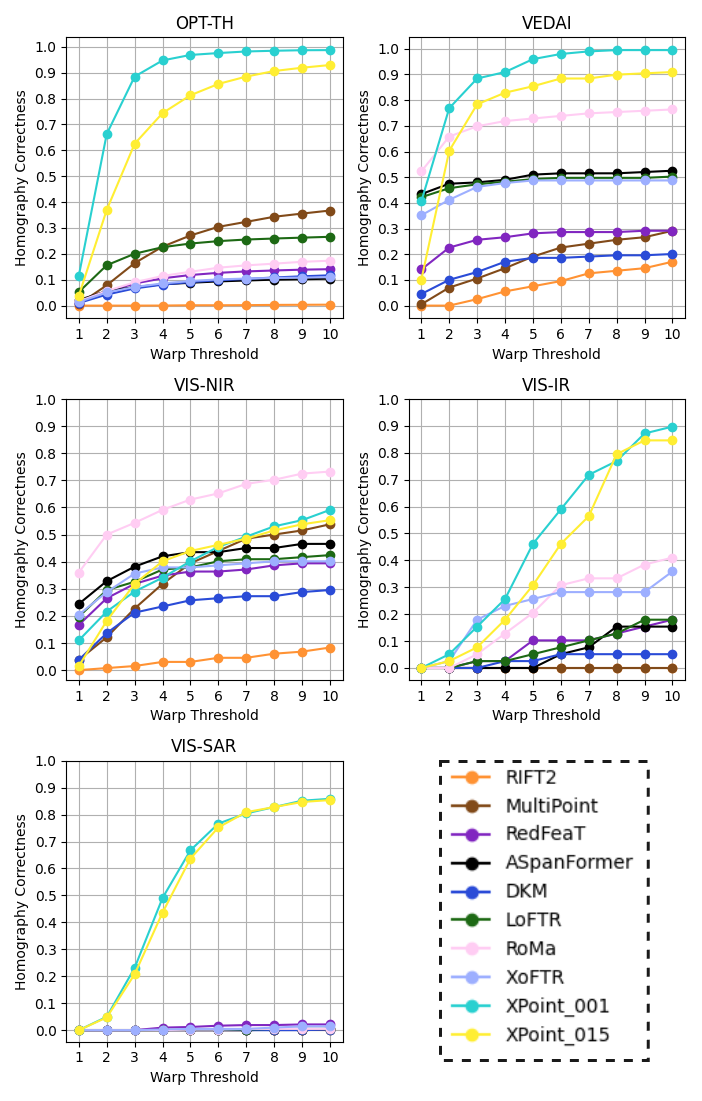}
    \caption{Homography Estimation results across datasets.}
    \label{fig:homography_results}
    \vspace{-0.5cm}
\end{figure}

While XPoint can estimate homography directly, simulations reveal that an indirect approach (Fig. \ref{fig:overall}(C)) using keypoints detected by our model yields more robust results. XPoint computes keypoints and descriptors, which can be matched using traditional methods like nearest neighbor search or deep learning-based approaches like LightGlue \cite{lightglue}. In this work,  we use the traditional pipeline for its simplicity and compatibility across benchmarks in detector-based methods, including our method. Outliers are removed using MAGSAC \cite{magsac} to estimate homography parameters. Benchmarks include detector-based methods, RIFT2\cite{rift2}, MultiPoint\cite{multipoint} and RedFeaT\cite{multimodal-lb3} and detector-free methods, ASpanFormer\cite{aspanformer}, DKM\cite{dkm}, LoFTR\cite{loftr}, RoMa\cite{roma} and XoFTR\cite{xoftr}. For our method, we test two variants, ``XPoint\_001'' and ``XPoint\_015'', which have keypoint  detection threshold values of 0.001 and 0.015, respectively. The evaluation results for all datasets are presented in Figures \ref{fig:rr_and_mscore} and \ref{fig:homography_results}.

\vspace{-0.3cm}
\subsection{Feature Matching Performance}
 Figure \ref{fig:rr_and_mscore} highlights the feature matching performance of detector-based methods, RIFT2\cite{rift2}, MultiPoint\cite{multipoint} and RedFeaT\cite{multimodal-lb3} and detector-free methods, ASpanFormer\cite{aspanformer}, DKM\cite{dkm}, LoFTR\cite{loftr}, RoMa\cite{roma} and XoFTR\cite{xoftr} and our proposed method's variants XPoint\_001 and XPoint\_015 on five different datasets. In this figure, Repeatability and Matching Score are selected as primary evaluation metrics and calculated at pixel threshold of 5.

\textbf{Repeatability.}
Figure \ref{fig:rr_and_mscore}, row ``Repeatability'', presents the Repeatability scores across various datasets, with each dataset occupying a separate column. Our proposed XPoint\_001 variant consistently outperforms established models such as RoMa \cite{roma} and RedFeaT \cite{multimodal-lb3} across all datasets. In contrast, the XPoint\_015 variant demonstrates slightly lower performance. These findings highlight the robustness of XPoint in ensuring reliable keypoint detection, even across diverse spectral domains. Notably, XPoint excels in identifying candidate keypoints suitable for matching, showcasing its ability to effectively adapt to the nonlinear radiation distortion (NRD) effects inherent to multispectral imagery.

\textbf{Matching Score.} 
Figure \ref{fig:rr_and_mscore}, row ``Matching Score'', showcases the Matching Scores across multiple datasets. Our XPoint\_001 variant consistently outperforms other methods with a significant margin across all datasets, except in the VIS-NIR setting. This superior performance underscores the robustness of our approach in identifying reliable matches, particularly in challenging multispectral scenarios such as Optical-Thermal and VIS-SAR. Both XPoint\_001 and XPoint\_015 variants demonstrate exceptional performance, surpassing other methods in most cases. This indicates that the repeatable keypoints detected by our method are also highly matchable, highlighting the strength of our descriptor representation.

In the VIS-NIR setting, however, we are surpassed by detector-free methods. This dataset includes a wide variety of scenes, such as countryside, fields, forests, indoor environments, mountains, historic buildings, streets, urban areas, and water bodies. Given the limited training data of only 345 image pairs, matching the performance of detector-free methods—trained on large datasets with extensive scene diversity similar to VIS-NIR—becomes challenging. Nonetheless, our approach still outperforms all detector-based methods, including RedFeaT, the previous state-of-the-art in this setting. This achievement highlights the competitive advantage of our method among detector-based approaches, even with limited training data.

\vspace{-0.2cm}
\subsection{Homography Estimation and Image Registration Performance}
Figure \ref{fig:homography_results} presents the Homography Estimation results across various datasets. Our model, particularly XPoint\_001, consistently outperforms other methods by a substantial margin across all datasets, with performance improving as the pixel error threshold is relaxed. The only exception is the VIS-NIR setting, where our method performs on par with detector-free methods, with RoMa slightly surpassing us. This can be attributed to the fact that detector-free methods are trained on large datasets with extensive scene diversity similar to the VIS-NIR dataset, which includes countryside, fields, forests, buildings, and other areas of interest.

\begin{figure*}[h]
    \centering
    \includegraphics[width=0.9\textwidth]{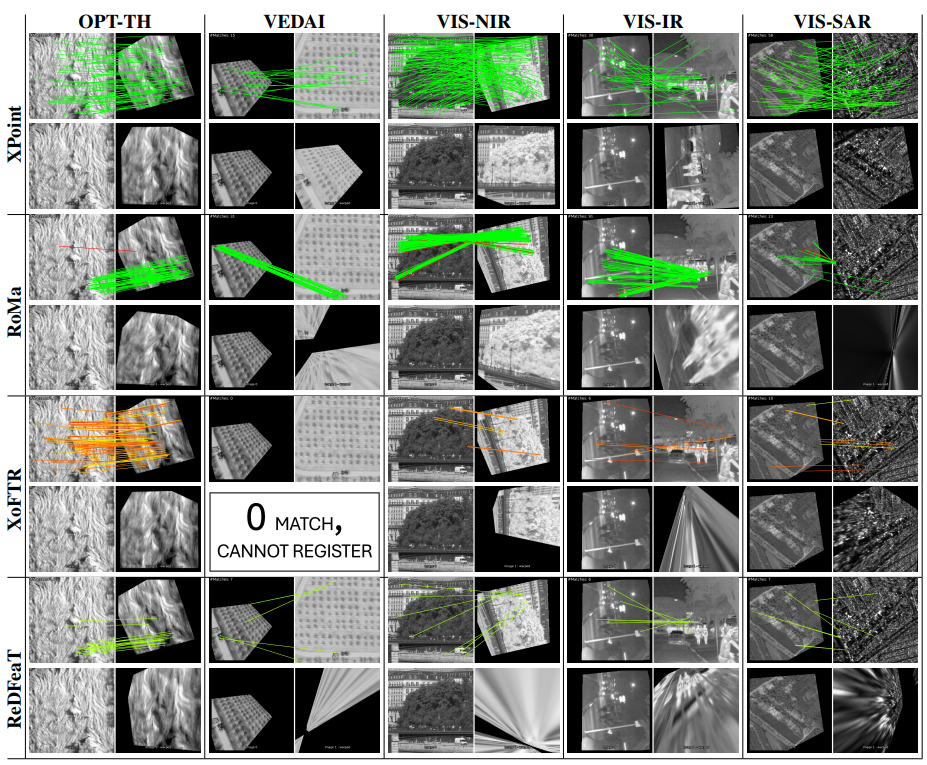}

    \caption{Qualitative results on five datasets—OPT-TH, VEDAI, VIS-NIR, VIS-IR, and VIS-SAR—using four methods: XPoint (ours), ReDFeat \cite{multimodal-lb3}, RoMa \cite{roma}, and XoFTR \cite{xoftr}. XPoint and ReDFeat are detector-based, while RoMa and XoFTR are detector-free. Each column shows a dataset, with two rows per method: the first row displays MAGSAC-filtered keypoint matches, and the second shows the result of image registration. Match confidence is indicated by a colormap from green (high) to red (low).}

    \label{fig:qualitative-results}
    \vspace{-0.6cm}
\end{figure*}

In datasets such as OPT-TH and VIS-SAR, however, our XPoint\_001 model achieves exceptional results, surpassing all other methods by a wide margin. This demonstrates the robustness of our approach in accurately estimating the homography matrix, a critical component for tasks like image registration. As shown in the qualitative results in Figure \ref{fig:qualitative-results}, our model registers images with exceptional precision, whereas competing methods often struggle due to poor matching performance. These results highlight the superior capability of XPoint not only in feature matching but also in achieving precise image registration.

\begin{figure}[h]
    \centering
    \includegraphics[width=\columnwidth]{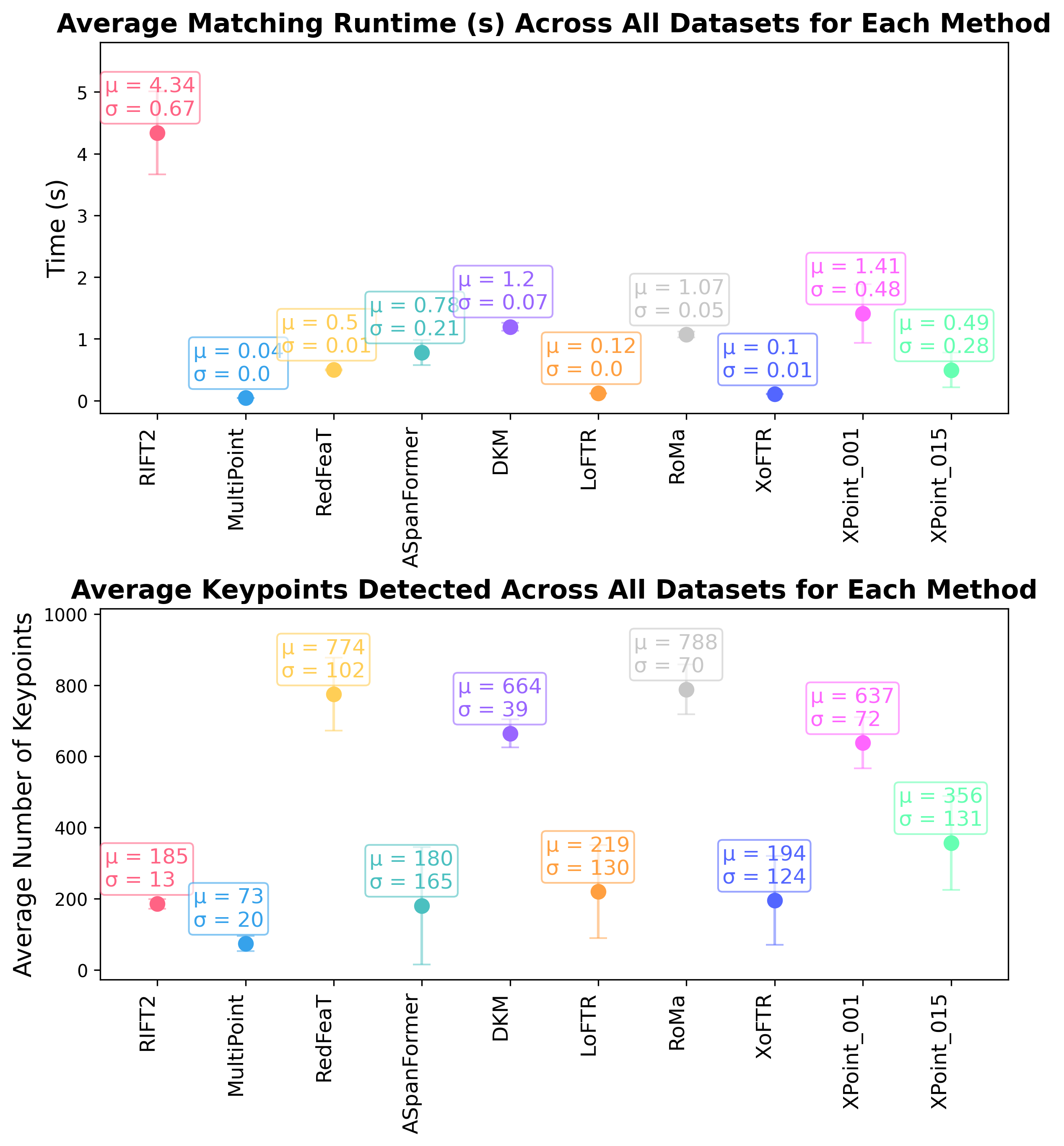}
    \caption{Comparison of average runtime (seconds) and keypoint counts across datasets for various methods.}
    \label{fig:time_error_plot}
\end{figure}

\vspace{-0.2cm}
\subsection{System Runtime and Keypoint Statistics}
We evaluate runtime and keypoint statistics for all methods, including XPoint variants, on an NVIDIA RTX 2080Ti GPU. Figure \ref{fig:time_error_plot} compares the average keypoints detected and runtime (seconds) across multiple datasets. XPoint\_001 detect an average of 637 keypoints in 1.41 seconds per image pair, while XPoint\_015 detect 356 keypoints in 0.49 seconds, reflecting a trade-off between keypoint density and processing speed. Compared to other methods, XPoint variants strike a favorable balance between runtime and detection. For instance, RoMa achieves high keypoint counts (788) but averages a longer runtime (1.07 seconds), while MultiPoint is faster (0.04 seconds) but detects fewer keypoints (73). This balance makes XPoint suitable for real-time applications, with configurable variants like XPoint\_001 and XPoint\_015 adapting to different precision or computational needs.

\vspace{-0.3cm}
\subsection{Ablation Study}

\subsubsection{Analysis of Architectural Modifications}

\begin{table}[h]
    \centering
    \caption{Ablation Study of XPoint Architectural Configurations on the Optical-Thermal Dataset.}
    \resizebox{\columnwidth}{!}{%
    \begin{tabular}{l c c c c}
        \toprule
        \multicolumn{1}{c}{} &
        \multicolumn{2}{c}{\textbf{Homographic Adaptation}} & \multicolumn{2}{c}{\textbf{Architecture}} \\
        \cmidrule(lr){2-3} \cmidrule(lr){4-5}
        \textbf{Configuration} & \textbf{Acceptance} & \textbf{Base Detector} & \textbf{Encoder} & \textbf{Homography Head} \\
        \midrule
        RIFT2 & - & - & - & - \\
        Base & Gaussian & SURF & CNN & \xmark \\
        XP1 & Windowing [26] & SURF & CNN & \xmark \\
        XP2 & Proposed W. & SURF & CNN & \xmark \\
        XP3 & Proposed W. & RIFT2 & CNN & \xmark \\
        
        XP4 & Gaussian & SURF & Swin-T & \xmark \\
        
        %XP5 & Product & SURF & Swin-T & \cmark \\
        
        XP5 & Windowing & SURF & Swin-T & \xmark \\
        XP6 & Windowing & SURF & Swin-T & \cmark \\
        XP7 & Proposed W. & RIFT2 & Swin-T & \cmark \\
        
        XPoint & Proposed W. & RIFT2 & VMamba-T & \cmark \\
        
        \midrule
        \multicolumn{5}{c}{\textbf{Performance Metrics}} \\
        \midrule
        \textbf{Method} & \textbf{Repeatibility} & \textbf{Matching Score} & $\varepsilon = 3$ & $\varepsilon = 5$ \\
        \midrule
        RIFT2 & 0.534 & 0.006 & 0 & 0.001 \\
        Base & 0.286 & 0.156 & 0.164 & 0.271 \\
        XP1 & 0.311 & 0.149 & 0.181 & 0.293 \\
        XP2 & 0.346 & 0.145 & 0.201 & 0.350 \\
        
        XP3 & 0.439 & 0.122 & 0.257 & 0.451 \\
        
        XP4 & 0.323 & 0.340 & 0.256 & 0.374 \\

        %XP5 & 0.316 & 0.335 & 0.245 & 0.352 \\
        
        XP5 & 0.379 & 0.344 & 0.296 & 0.482 \\
        XP6 & 0.391 & 0.373 & 0.344 & 0.521 \\
        
        XP7 & 0.462 & 0.401 & 0.573 & 0.794 \\
        
        XPoint & \textbf{0.469} & \textbf{0.426} & \textbf{0.631} & \textbf{0.813} \\
        \bottomrule
    \end{tabular}%
    }
    \vspace{-0.2cm}
    \label{tab:ablation-new}
\end{table}

We conducted an ablation study on the Optical-Thermal dataset to assess the impact of various components within the XPoint architecture (see Table~\ref{tab:ablation-new}). The performance metrics evaluated include Repeatability, Matching Score, and Homography Estimation accuracy at pixel errors $\varepsilon = 3$ and $\varepsilon = 5$. In all experiments, the keypoint detection threshold is set to 0.015. In Table \ref{tab:ablation-new}, \textbf{RIFT2} evaluates the RIFT2 model directly as a keypoint and descriptor detection method.  The \textbf{Base Model} acts as our initial baseline, using SURF as the base detector and a Gaussian for keypoint acceptance during homographic adaptation. It includes a CNN encoder but does not use a homography head. The XP configurations, labeled from \textit{XP1} to \textit{XPoint}, progressively add architectural components, leading up to the final XPoint model.

% \textbf{Baseline and Naming Convention:}

% \begin{itemize} \item \textbf{RIFT2:} This configuration evaluates the RIFT2 model directly as a keypoint and descriptor detection method. \item \textbf{Base Model:} Serves as our initial baseline, using the SURF detector with the Gaussian product rule during homographic adaptation, paired with a CNN encoder, and without the homography head. \item \textbf{XP Configurations:} Configurations labeled \textit{XP1} to \textit{XPoint} progressively incorporate additional architectural components, culminating in the final XPoint model. \end{itemize}

\textbf{RIFT2 vs. Base Model:} The \textit{RIFT2} configuration demonstrates a clear advantage in Repeatability, achieving a higher score (0.534) compared to the \textit{Base} model (0.286), which indicates its strength in detecting repeatable keypoints. However, this comes with trade-offs: RIFT2's low Matching Score (0.006 vs. 0.156 for \textit{Base}) and minimal homography accuracy underscore the limitations of handcrafted descriptors in multispectral matching. This suggests that while RIFT2 excels in keypoint detection, it struggles with accurately matching descriptors across different modalities due to poor descriptor quality.

\textbf{Impact of Keypoint Acceptance Techniques on Keypoint Detection:} Introducing the original windowing method in \textit{XP1} enhances Repeatability from 0.286 to 0.311 and improves homography estimation accuracy. The slight drop in Matching Score indicates that although more repeatable keypoints are detected, they may be less uniquely described due to the limitations of the CNN encoder.

Implementing the proposed windowing technique in \textit{XP2} further increases Repeatability to 0.346 and improves homography estimation. The continued slight impact on Matching Score reflects challenges in extracting distinctive descriptors as the number of keypoints rises.

\textbf{Effectiveness of the Base Detector:}  Building on the insights from RIFT2’s keypoint detection, replacing the SURF detector with RIFT2 in the \textit{XP3} experiment leads to notable improvements. The Repeatability increases to 0.439, and homography estimation scores show a significant boost. However, the decline in Matching Score (to 0.122) reflects the challenges posed by higher keypoint density, as the CNN encoder faces difficulty in extracting distinctive descriptors in this context. This finding reinforces the role of RIFT2 as a powerful base detector for enhancing Repeatability and homography accuracy when combined with a learning-based approach, such as a CNN. By utilizing RIFT2’s robust keypoint generation alongside a CNN-based descriptor matching, this setup surpasses the limitations observed in the standalone \textit{RIFT2} experiment.

\textbf{Impact of Encoder Architecture:} Upgrading to a Swin Transformer encoder in \textit{XP4} enhances all metrics, with Matching Score notably increasing to 0.340. This demonstrates the Swin-T encoder's superior ability to extract distinct and robust descriptors. Reintroducing windowing in \textit{XP5} further improves performance. The Repeatability increases to 0.379, and homography estimation scores improve at both $\varepsilon = 3$ and $\varepsilon = 5$. These results show that the Swin-T encoder effectively handles an increased number of keypoints, leading to more robust matching.

\textbf{Contribution of the Homography Head:} Adding a homography head in \textit{XP6} provides a boost across all metrics, raising Repeatability to 0.391 and Matching Score to 0.373. The improved homography scores confirm the homography head’s role in enhancing geometric consistency and refining keypoints for more accurate alignment.

\textbf{Synergistic Effects of Combined Components:} Combining RIFT2 as the base detector, the proposed windowing method, the Swin-T encoder, and the homography head in \textit{XP7} leads to substantial gains. The Repeatability reaches 0.462, Matching Score increases to 0.401, and homography estimation scores reach 0.573 and 0.794 at $\varepsilon = 3$ and $\varepsilon = 5$, respectively. This configuration showcases the complementary benefits of these components in enhancing keypoint learning and matching.

\textbf{Superior Performance with VMamba-T Encoder:} The final \textit{XPoint} model incorporates the VMamba-T encoder, achieving the highest metrics: an Repeatability of 0.469, Matching Score of 0.426, and homography estimation scores of 0.631 and 0.813. This underscores the VMamba-T encoder's superior capability in cross-modal matching through advanced feature extraction.

In summary, this ablation study highlights the incremental impact of each component within the XPoint architecture. 

\textbf{Advanced keypoint acceptance methods} enhance keypoint repeatability, providing more consistent detection across varying conditions.

\textbf{Robust base detectors}, such as RIFT2, improve keypoint detection in challenging cross-modal scenarios, addressing variations in spectrum and viewpoint.

\textbf{Pretrained Encoder architectures}, like Swin-T and VMamba-T, increase descriptor distinctiveness and matching quality, capturing richer and more reliable features.

\textbf{Homography Head} further refines keypoints, ensuring geometric consistency and enhancing homography estimation accuracy. 

Together, these components contribute significantly to improvements in Repeatability, Matching Scores, and homography estimation, validating the effectiveness of the XPoint architecture.

\begin{figure}[htp]
    \centering
    \resizebox{\columnwidth}{!}{%
    \begin{tabular}{c  c | c}
        & \textbf{SURF} & \textbf{RIFT2} \\ [-0.5cm] 
        
        \rotatebox{90}{\textbf{Gaussian}} & 
        \stackunder[5pt]{\includegraphics[width=0.45\linewidth]{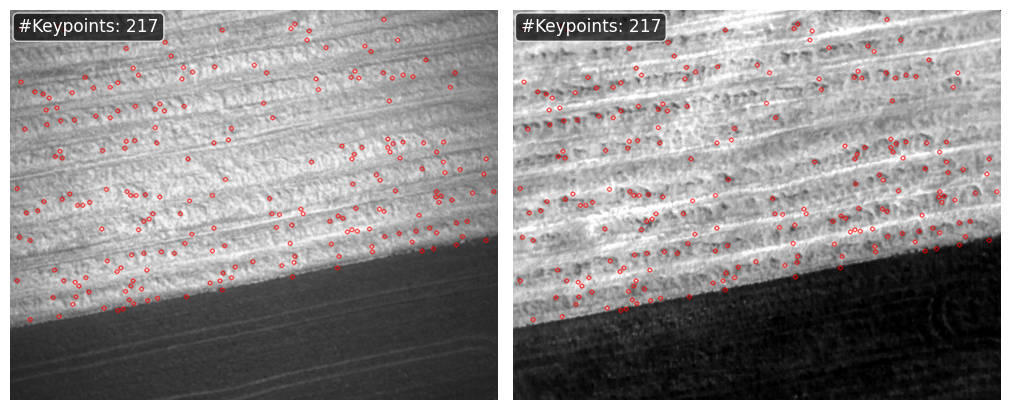}}{Nkp: 217, 217} &
        \stackunder[5pt]{\includegraphics[width=0.45\linewidth]{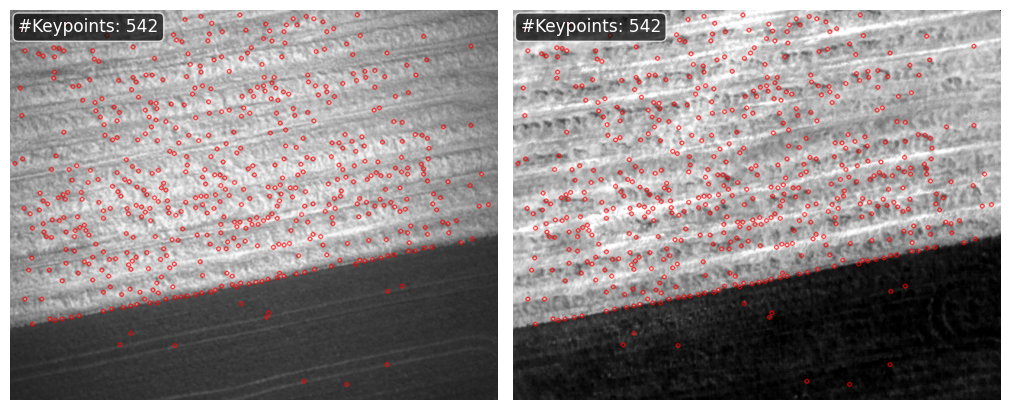}}{Nkp: 542 , 542} \\[0.05cm]
        
        \rotatebox{90}{\textbf{Windowing}} & 
        \stackunder[5pt]{\includegraphics[width=0.45\linewidth]{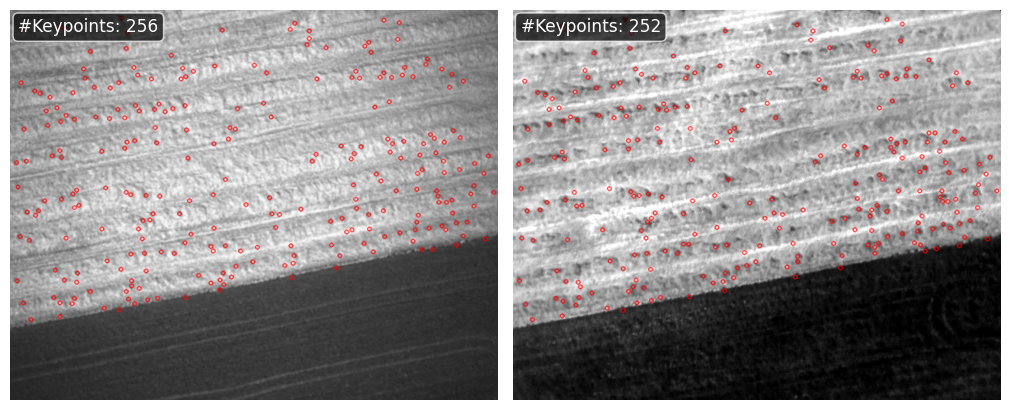}}{Nkp: 256 , 252} & 
        \stackunder[5pt]{\includegraphics[width=0.45\linewidth]{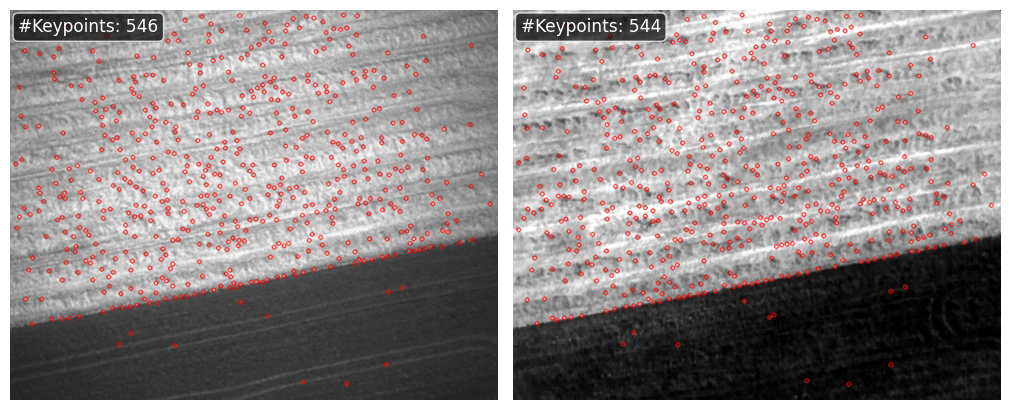}}{Nkp: 546 , 544} \\[-0.2cm]
        
        \rotatebox{90}{\textbf{Proposed W.}} & 
        \stackunder[5pt]{\includegraphics[width=0.45\linewidth]{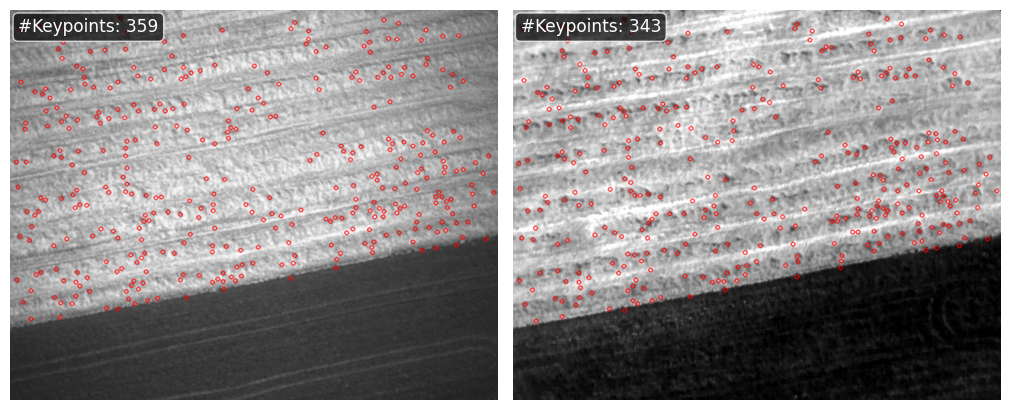}}{Nkp: 359 , 343} & 
        \stackunder[5pt]{\includegraphics[width=0.45\linewidth]{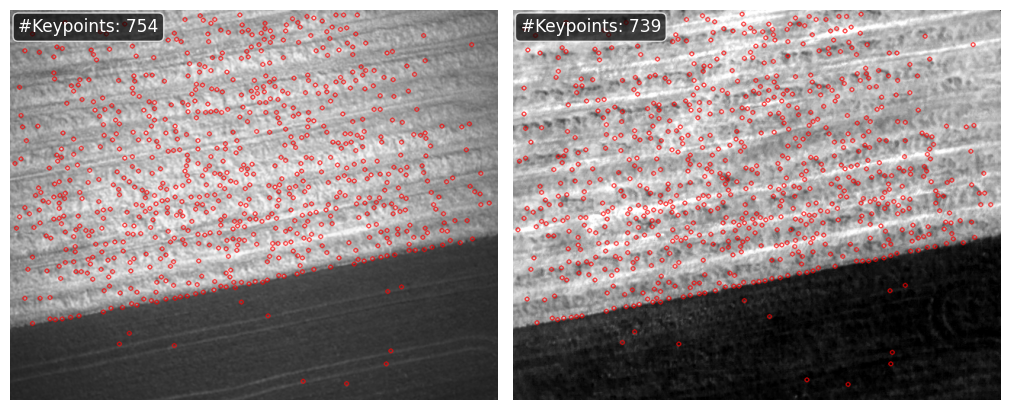}}{Nkp: 754 , 739 } \\ 
    \end{tabular}
    }
    \caption{Visual comparison of keypoint detection results for optical-thermal image pairs using different base detectors and homographic adaptation methods (Nkp: number of keypoints).}
    \label{tab:ablation-homographic-adaptation}
    \vspace{-0.3cm}
\end{figure}

\subsubsection{Effect of Base Detector and Proposed Windowing Technique in Homographic Adaptation Stage}

This section evaluates the impact of different base detectors and keypoint acceptance techniques on keypoint detection in multispectral pairs, comparing SURF and RIFT2 detectors with three adaptation methods: Gaussian, Windowing, and our Proposed windowing technique (see Table~\ref{tab:ablation-homographic-adaptation}). The \textbf{Gaussian} method generates a keypoint heatmap by performing element-wise multiplication on image pairs, followed by smoothing with a Gaussian filter. However, it lacks adaptability to spectral variations. The \textbf{Windowing} technique~\cite{windowing} improves flexibility by matching keypoints within a defined window, but applies a binary acceptance criterion. Our \textbf{Proposed} windowing technique further refines this by treating keypoints probabilistically and accumulating observations across homographies, resulting in more robust keypoint detection adaptable to both spectral and viewpoint changes. Among detectors, \textbf{RIFT2} outperforms SURF across all adaptation methods, owing to its ability to address nonlinear radiation distortion (NRD) and reduce spectral domain gaps. Combining RIFT2 with our proposed technique produces a dense, reliable keypoint set, proving optimal for cross-spectral matching. In summary, this pairing enhances keypoint density and distribution across multispectral images, effectively handling spectral and viewpoint variability for improved cross-spectral performance.

\begin{figure}[htp]
    \centering
    \resizebox{\columnwidth}{!}{%
    \begin{tabular}{c  c | c}
        & \textbf{Cross-Entropy} & \textbf{Weighted Cross-Entropy} \\ [0cm] 
        \rotatebox{90}{\textbf{Example 1}} & 
        \includegraphics[width=0.45\linewidth]{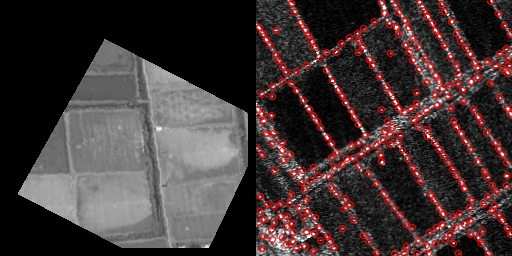} & 
        \includegraphics[width=0.45\linewidth]{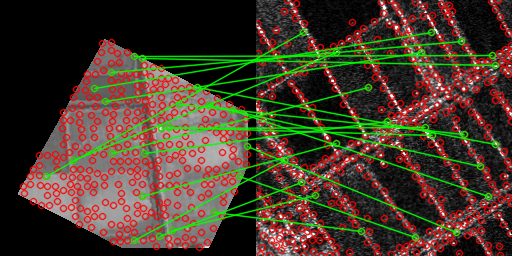} \\[0.1cm] 
        
        \rotatebox{90}{\textbf{Example 2}} & 
        \includegraphics[width=0.45\linewidth]{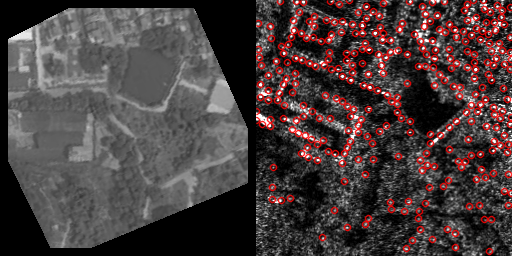} & 
        \includegraphics[width=0.45\linewidth]{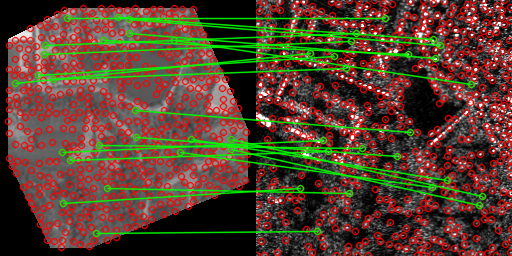} \\[0.1cm] 
        
    \end{tabular}
    }
    \caption{Visual comparison between Cross-Entropy and Weighted Cross-Entropy loss strategies.}
    \label{tab:ablation-class-imbalance}
    \vspace{-0.2cm}
\end{figure}

\subsubsection{Handling Keypoint Sparsity Problem using Weighted Cross-Entropy Loss}

In the experiments shown in Table~\ref{tab:ablation-class-imbalance}, two examples are presented in each row, with two loss functions compared in the columns: regular Cross-Entropy (left) and Weighted Cross-Entropy (right). VIS-SAR, trained with standard cross-entropy, struggles with class imbalance, as evident in the images where one spectrum is densely filled with keypoints while the other contains none, resulting in zero matches between the two modalities. This imbalance limits the model’s ability to predict accurate correspondences between the spectra, diminishing the matching score. To address this issue, a weighted cross-entropy loss was employed during training. This approach assigns lower weights to the losses of overrepresented classes, such as the dustbin class in the VIS-SAR setting, promoting more balanced predictions across both spectra, as demonstrated in the examples. The result is an improvement in keypoint distribution and matching accuracy between the modalities. Weighted loss mitigates the negative impact of class imbalance and increases the robustness of the model in cross-spectral matching tasks.

\section{Conclusion And Future Work}
This paper introduces XPoint, a self-supervised framework for multispectral image matching with improved ground truth label extraction stage, transfer learning from VSS-based VMamba encoders, homography regression head as a geometric constraint, and flexible detector loss. Our evaluations on five different datasets demonstrate its superior performance in feature matching and homography estimation metrics. In the future, we will focus on enhancing computational efficiency through lighter encoders, refining decoder designs for more accurate keypoint analysis, and improving the homography regression head for seamless integration with the matching process. These updates aim to elevate our model's performance and adaptability for diverse applications. 

\section*{Acknowledgments}
This work is supported in part by TUBITAK Grant - Project No: 118E891.

\vspace{-0.2cm}

\bibliography{bibliography}% common bib file
\bibliographystyle{IEEEtran}

\end{document}